\pdfoutput=1

\documentclass[11pt]{article}

\usepackage{EACL2023}

\usepackage{times}
\usepackage{latexsym}

\usepackage[T1]{fontenc}

\usepackage[utf8]{inputenc}

\usepackage{microtype}

\usepackage{enumitem,kantlipsum}

\usepackage{xspace}
\usepackage{amsmath,amssymb}
\usepackage{xcolor}
\usepackage{graphicx}

\usepackage{array,multirow}
\usepackage{booktabs}
\newcommand{\ra}[1]{\renewcommand{\arraystretch}{#1}}

\usepackage{bm}
\usepackage{subfig}
\usepackage{wrapfig}

\definecolor{xsumcolor}{HTML}{a6611a}
\definecolor{cnndmcolor}{HTML}{dfc27d}
\definecolor{mn500color}{HTML}{018571}
\definecolor{mn800color}{HTML}{80cdc1}

\definecolor{mintgreen}{rgb}{0.15, 0.75, 0.50}

\newcommand\mint{\textsc{Mint}\xspace }

\newcommand\bleu{\textsc{Bleu}\xspace }
\newcommand\rouge{\textsc{Rouge}\xspace }

\newcommand\ngram{\textit{n}-gram\xspace }
\newcommand\ngrams{\textit{n}-grams\xspace }
\renewcommand\vec[1]{\ensuremath\bm{#1}}
\newcommand\fragments{\mathcal{F}(\vec{x},\vec{y})}
\newcommand\facth{\textsc{FactH}\xspace}
\newcommand\muFactH{$\mu$\textsc{FactH}\xspace}

\newcommand\bart{\textsc{Bart}\xspace}

\renewcommand\eqref[1]{Equation~\ref{eq:#1}}
\newcommand\xfigref[1]{Figure~\ref{fig:#1}}
\newcommand\fref[1]{Fig.~\ref{fig:#1}}
\newcommand\secref[1]{Section~\ref{sec:#1}}
\newcommand\sref[1]{Sec.~\ref{sec:#1}}
\newcommand\appref[1]{Appendix~\ref{appendix:#1}}
\newcommand\tabref[1]{Table~\ref{tab:#1}}

\usepackage{amsmath}
\DeclareMathOperator*{\argmax}{arg\,max}

\newcommand\mucell[2]{
  \textcolor{mintgreen}{#1}\ \ \small #2}

\newcommand{\note}[3]{}

\title{Evaluating the Tradeoff Between Abstractiveness and Factuality \\
in Abstractive Summarization}

\author{
  Markus Dreyer$^1$ \quad
  Mengwen Liu$^1$ \quad
  Feng Nan$^1$ \quad
  Sandeep Atluri$^1$ \quad
  Sujith Ravi$^2$\thanks{$^\ast$Work conducted during his position at Amazon.} \quad
  \\
  Amazon$^1$ \qquad SliceX$^2$ AI\\
  \texttt{\{mddreyer, mengwliu, nanfen, satluri\}@amazon.com}\\
  \texttt{ravi.sujith@gmail.com}}

\begin{document}
\maketitle
\begin{abstract}

Neural models for abstractive summarization tend to generate output
that is fluent and well-formed but lacks semantic faithfulness, or
factuality, with respect to the input documents. In this paper, we
analyze the tradeoff between abstractiveness and factuality of
generated summaries across multiple datasets and models, using
extensive human evaluations of factuality. In our analysis, we
visualize the rates of change in factuality as we gradually increase
abstractiveness using a decoding constraint, and we observe that,
while increased abstractiveness generally leads to a drop in
factuality, the rate of factuality decay depends on factors such as
the data that the system was trained on.  We introduce two datasets
with human factuality judgements; one containing 10.2k
generated summaries with systematically varied degrees of
abstractiveness; the other containing 4.2k summaries from five
different summarization models.  We propose new factuality metrics
that adjust for the degree of abstractiveness, and we use them to
compare the abstractiveness-adjusted factuality of previous
summarization works, providing baselines for future
work.\footnote{Code and data are available
at \url{https://github.com/amazon-science/abstractive-factual-tradeoff}.}

\end{abstract}

\section{Introduction}\label{sec:intro}

Summarization is the task of generating a semantically faithful,
well-formed and concise text representation of the input.
Automatically generated summaries have traditionally
been \textit{extractive}
\cite{Luhn1958TheAC,Edmundson1969NewMI,10.5555/645853.669480,10.5555/1622487.1622501,
  wong-etal-2008-extractive}, leading to issues with readability and
coherence, as different extracted fragments may not fit well when
taken out of their original contexts \cite{Poibeau2012}. Researchers
have also invested in methods for
\textit{abstractive} summarization, aiming to paraphrase the input
documents' main points without borrowing their exact lexical
expressions \cite{radev-mckeown-1998-generating,Saggion2002GeneratingIS,Ganesan2010OpinosisAG,Genest2012FullyAA,radford2019language,
gehrmann-etal-2019-generating, lewis2019bart, pmlr-v119-zhang20ae}.
Abstractive summaries generated by today's neural models tend to be
fluent and well-formed, but lack semantic faithfulness
\cite{Cao2018faithful,kryscinski-etal-2019-neural}. Observed rates of
factual errors in abstractive summaries have ranged from 30\% to over
75\% \cite{Cao2018faithful,maynez-etal-2020-faithfulness}.  The
research community is developing automatic factuality metrics
\cite{wang2020asking,kryscinski2020evaluating,Goodrich2019assessing,goyal-durrett-2020-evaluating,ribeiro-etal-2022-factgraph}
and methods that attempt to increase factuality
\cite{Fan2018robust,scialom-2019-answers-unite,Zhang2019optimizing,Falke2020ranking,cao-wang-2021-cliff}.
\begin{figure}[t] 
  \centering \includegraphics[width=\columnwidth]{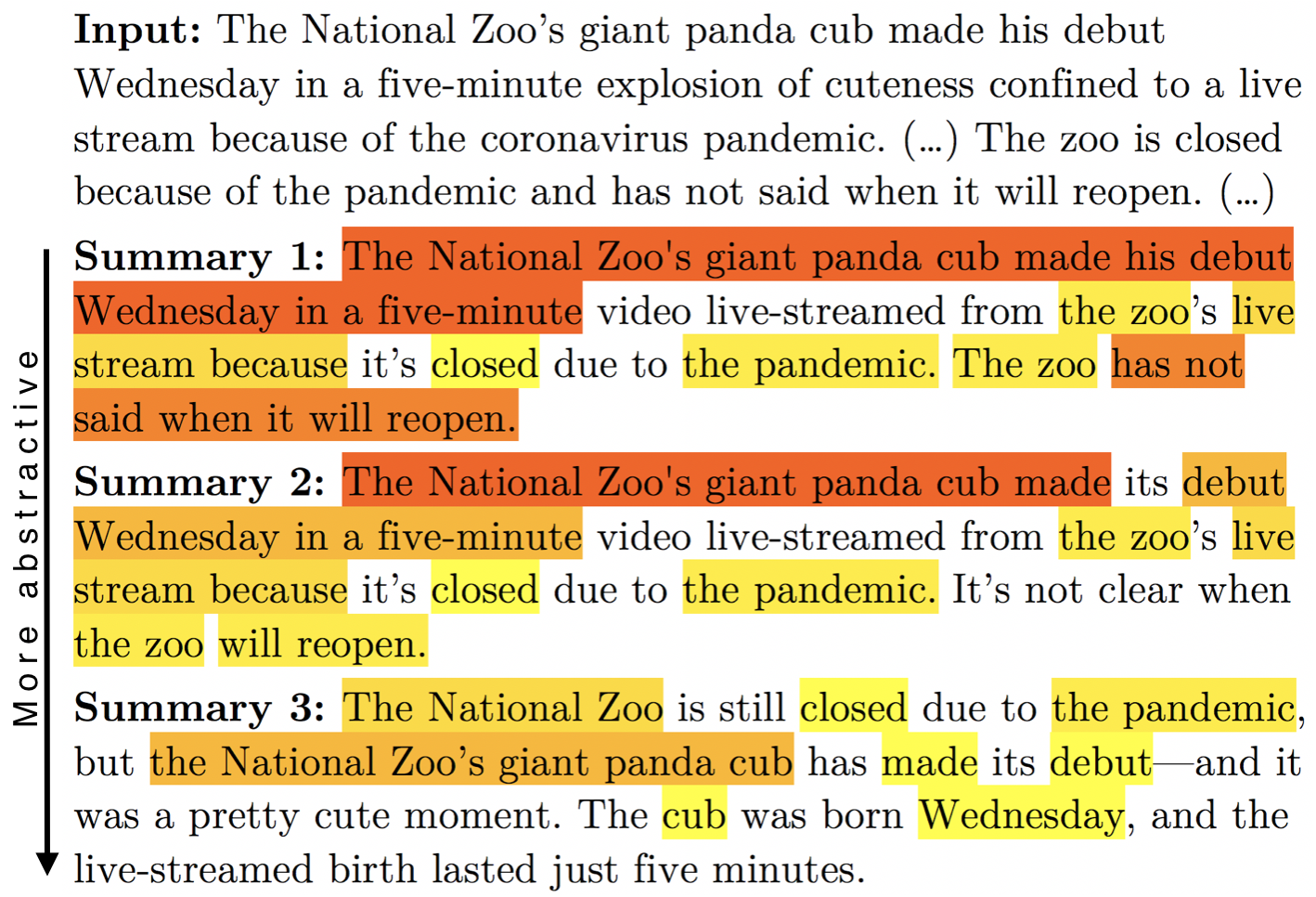} \caption{Three
  successively more abstractive summaries generated from the same
  input article, with \mint abstractiveness scores
  (\secref{mintscore}) of 46.1\%, 67.2\%, 79.5\%. Fragments extracted
  from the input are marked from red (longer fragments) to yellow
  (shorter fragments). The bottom summary has factual
  errors.}  \label{fig:more-abstractive}
\end{figure}
  However, the factuality problem of abstractive
  summaries cannot be well understood without considering
  the \textit{degree} of abstractiveness of a given summary:
{\parfillskip=0pt\parskip=0pt\par}%
\begin{wrapfigure}{L}{0.5\columnwidth}%
  \centering%
  \includegraphics[width=.5\columnwidth]{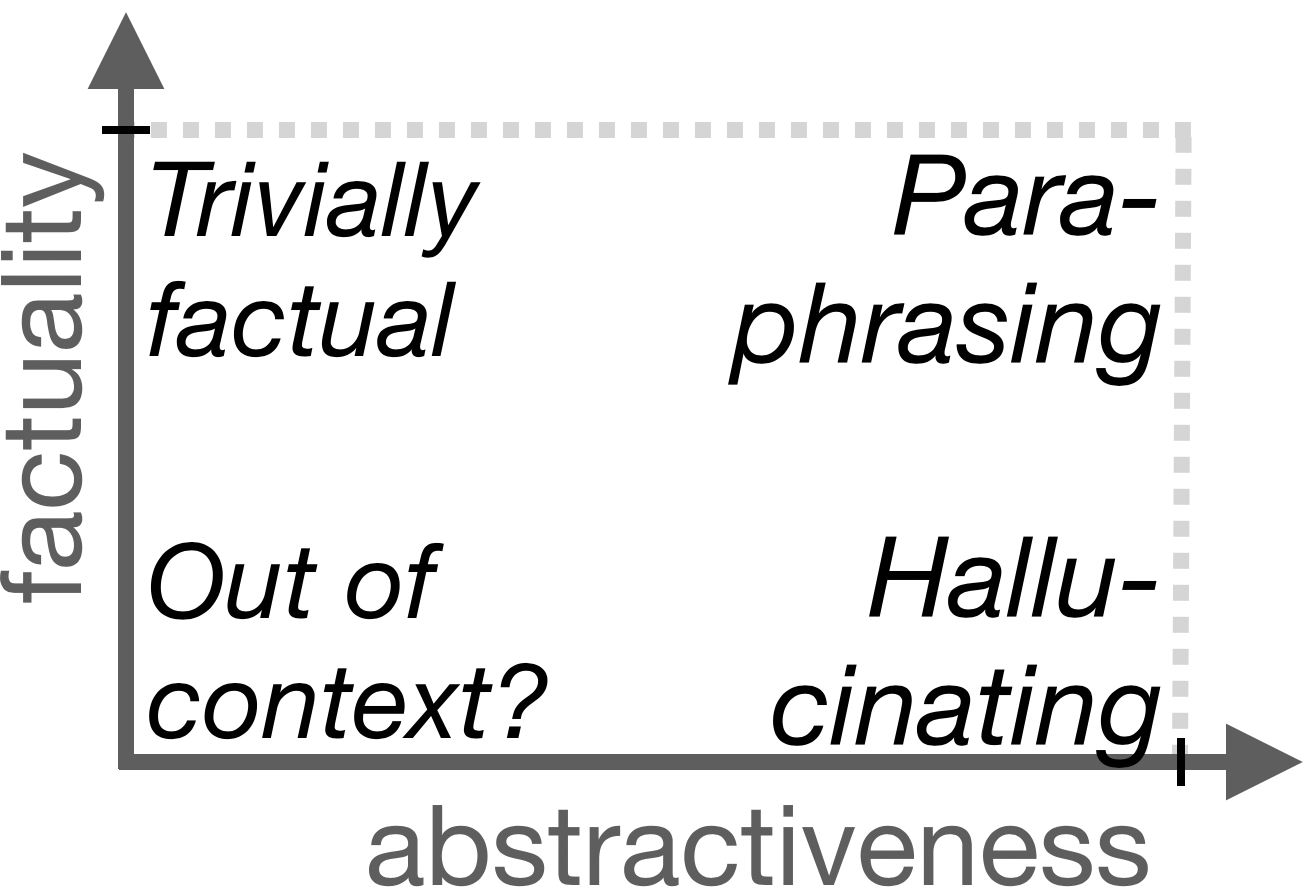}%
  \caption{Four extremes at the abstractiveness-factuality spectrum.}%
  \label{fig:four-corners}%
\end{wrapfigure}%
\noindent%
Any  summary is on a spectrum between \textit{extractive}
  and
\textit{abstractive}
\cite{see2017get}. Summaries that are extractive to a larger extent tend to be more factual since
copying text from the input into the summary rarely introduces factual
errors while the task of paraphrasing, which results in summaries that
are more \textit{abstractive}, is harder and prone to semantic errors.
As an example, \xfigref{more-abstractive} shows part of a Washington
Post article and three summaries with increasing abstractiveness,
which we have generated using our abstractiveness constraints
(\secref{slider}). The first two summaries are correct, but the third,
most abstractive, summary has factual errors, misinterpreting the
input.

Few authors have discussed this connection
explicitly. \newcite{lebanoff-etal-2019-analyzing} observe that
abstractive summaries consisting of concatenated extracted fragments
tend to be more factual than those created by more complex
fusion. \newcite{durmus2020feqa} observe that models trained on the
more \textit{extractive} CNN/DM dataset \cite{hermann2015teaching}
create more factual summaries than models trained on the more
\textit{abstractive} XSum dataset \cite{narayan-etal-2018-dont}. We
show that such models differ in factuality even when we bias them to
generate summaries that have similar levels of abstractiveness.  Our
analysis (\secref{tradeoff}) situates summarization models on the
spectrum outlined in \xfigref{four-corners}, where factual summaries
range from ``trivially factual'' (extractive) to truly
``paraphrasing'' (abstractive).  We make the following contributions:

\vspace{-3mm}
\begin{enumerate} %
\itemsep-1mm 

\item We  systematically explore the relationship of abstractiveness and
  factuality and show how factuality decays with increasing
  abstractiveness. We argue that factuality rates of different systems
  cannot be compared without taking their degrees of abstractiveness
  into account.

\item We introduce new factuality metrics that take abstractiveness
  into account and evaluate the abstractiveness-factuality tradeoff
  across various datasets and summarization models. We establish
  baselines that will allow others to demonstrate progress on
  mitigating the abstractiveness-factuality tradeoff.

\item We introduce a new dataset containing 10.2k summaries with
  systematically varied degrees of abstractiveness along with human
  factuality judgements, and a second dataset containing 4.2k
  summaries from five summarization models with their human factuality
  judgements.

\end{enumerate}

\section{Abstractiveness}\label{sec:abstractiveness}

\subsection{Measuring Abstractiveness}\label{sec:mintscore}

In this paper, we wish to analyze the relationship of abstractiveness
and factuality of generated summaries. We start by proposing a
comprehensive abstractiveness metric.  Abstractiveness measures the
amount of rephrasing, i.e., the degree to which the words, phrases and
sequences of the generated text have \textit{not} been extracted from
the corresponding input; a fully abstractive summary method expresses
the main points of the input in its own words.  To measure
abstractiveness, most authors list the proportions of summary \ngrams
of varying lengths that are novel, i.e., do not occur in the
corresponding inputs
\cite{see2017get,narayan-etal-2018-dont,gao-etal-2019-write}.
\newcite{grusky-etal-2018-newsroom} proposed a new metric also
based on contiguous overlapping text spans, \textit{density},
measuring the average length of extracted fragments in a
summary. Others have proposed metrics that take
common \textit{non-contiguous} subsequences into account, e.g.,
\textit{perfect fusion$_k$} \cite{durmus2020feqa} measures the
percentage of summary sentences that assemble substrings from $k$
source sentences in their original order.

\begin{figure}
  \centering
  \includegraphics[width=.85\columnwidth]{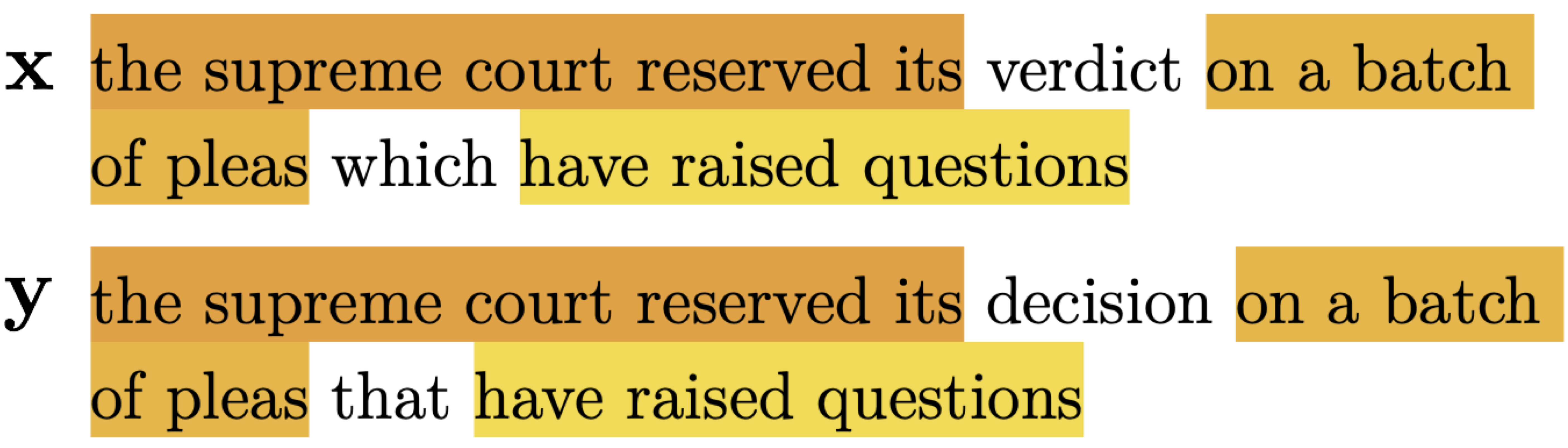}
  \caption{Example of input and highly extractive generated
    output. The color coding is the same as in \fref{more-abstractive}.}\label{fig:fragments-example}
\end{figure}

Based on these previous works, we define a comprehensive
abstractiveness metric that combines measures of contiguous and
non-contiguous extractive summary fragments, making it sensitive to
different kinds of abstractiveness and therefore suitable as a general
abstractiveness metric.  We define this metric as a ratio, in order to
facilitate combining it with a factuality metric of the same [0,1]
range (\secref{tradeoff}).  Let
$\chi(\vec{x},\vec{y})=\mathrm{hmean}(p_1, p_2, p_3, p_4,
\mathrm{lcsr})$ be a measure of \textit{extractive} overlap between input
$\vec{x}$ and summary $\vec{y}$, using the harmonic mean of multiple
component measures. Each $p_n$, short for $p_n(\vec{x}, \vec{y})$, is
the \ngram precision of the \ngrams in $\vec{y}$ with respect to
$\vec{x}$, i.e., the percentage of \ngrams in $\vec{y}$ that are
extracted from $\vec{x}$.\footnote{We smooth all \ngram
counts \cite{chen-cherry-2014-bleu} to avoid undefined or zero
harmonic mean values in highly abstractive
summaries. See \appref{mint} for details.} Following common
practice \cite{papineni-etal-2002-bleu}, we use \ngrams up to length
four. We do not include density in $\chi(\vec{x},\vec{y})$ as its 
range is unbounded. The measure $\mathrm{lcsr}$ (longest
common subsequence ratio), short for $\mathrm{lcsr}(\vec{x},\vec{y})$,
is the length of the longest common subsequence (LCS) between
$\vec{x}$ and $\vec{y}$ divided by the length of
$\vec{y}$. $\mathrm{lcsr}$, inspired by \rouge-L
\cite{lin-2004-rouge}, generalizes perfect fusion$_k$ to consider
\textit{all} instances of non-contiguous overlaps between input and
summary. Adding a measure of non-contiguous overlap is important as it
detects overlaps that are long but broken up by minor changes, such
as synonyms, as in the example in \xfigref{fragments-example}. Finally,
the \mint ({\bf M}etric for lexical {\bf in}dependence of generated
{\bf t}ext) abstractiveness measure is defined as
$\mint(\vec{x},\vec{y})=1-\chi(\vec{x},\vec{y})$. For a set of inputs
and their summaries, we report the average \mint score. See
\xfigref{more-abstractive} for the \mint scores of three increasingly
abstractive example summaries. In \secref{experiments}, we show
that \mint scores correlate hightly with density
scores. %

The described \mint score capitalizes on prior work to provide a
comprehensive and unified metric for abstractiveness of conditionally
generated text, combining measures of contiguous and non-contiguous
overlap into a single percentage score. The implementation of \mint we
provide will facilitate standardized comparisons of abstractiveness
across different works.

\subsection{Nonlinear Abstractiveness Constraints}\label{sec:slider}

We now introduce nonlinear abstractiveness constraints (NAC), which
enable us to control the degree of abstractiveness at decoding time;
it will allow us to use a trained summarization model to decode input
multiple times while applying constraints to control the
abstractiveness of the generated text output (e.g.,
see \xfigref{more-abstractive}). We will use this technique to analyze
the impact of abstractiveness on factuality (\secref{tradeoff}).

Let $\fragments$ be the set of the longest extractive fragments in the
decoding output $\vec{y}$ with respect to the input
$\vec{x}$. In \xfigref{more-abstractive}, such fragments are marked in
color for each summary.  We define a function $\lambda_h(|\vec{f}|)$
that assigns a discount probability to any extractive fragment
$\vec{f}\in\fragments$:

\begin{equation} \label{eq:lambda}
\lambda_{h}(|\vec{f}|) = 2^{-|\vec{f}|^2/h^2}
\end{equation}

\begin{figure}[t] 
  \centering
  \includegraphics[width=.8\columnwidth]{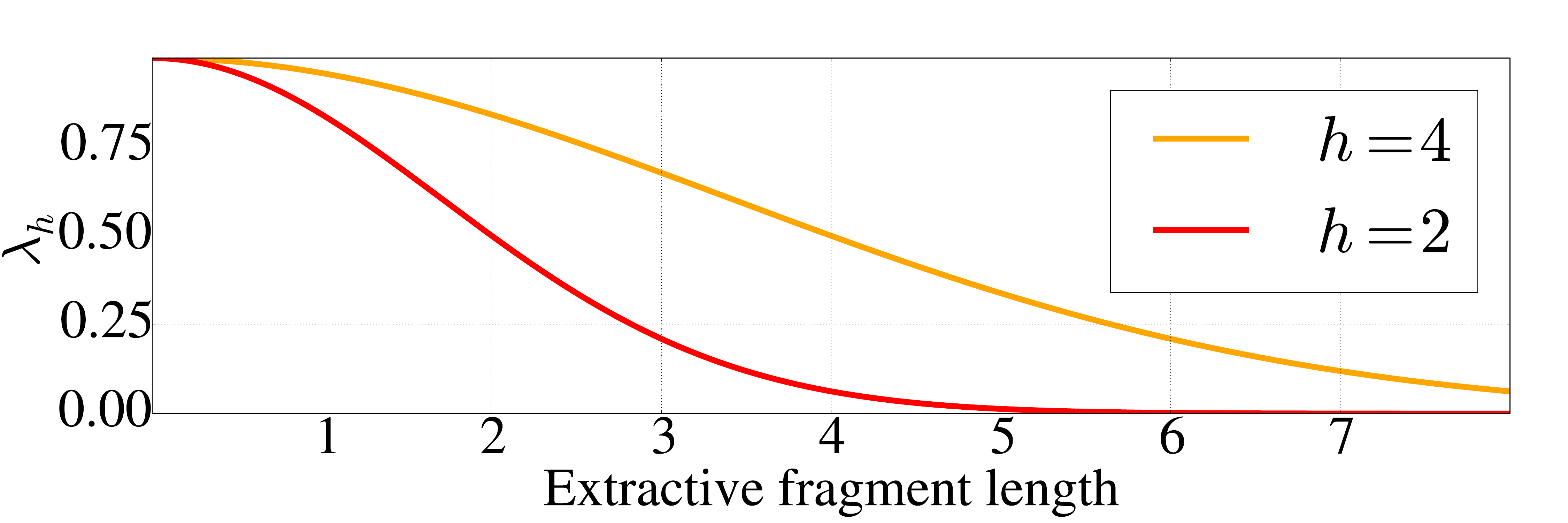}
  \caption{$\lambda_h$ defines discounts for extractive fragments based on
    their lengths. Smaller $h$ values lead to more abstractive summaries.}
  \label{fig:lambda2}
\end{figure}

We configure this function\footnote{Additionally, the exponent used in
$|\vec{f}|^2$ and $h^2$ could be configured, but we keep it at $2$ in
our experiments. A larger exponent would result in a steeper descent
around $h$.} with $h$, interpreted as the length of an extracted
fragment for which $\lambda_h=0.5$.
Decreasing $h$ results in a $\lambda_h$ that discounts shorter extractive fragments
more strongly, leading to increased abstractiveness (see \xfigref{lambda2}).
Our discount penalty grows nonlinearly, affecting longer extractive
fragments more strongly than multiple shorter ones with the same
combined length. To see why we choose a \textbf{nonlinear penalty},
consider for example that extracting a 10-gram makes a summary more
extractive than using ten words from the article separately, since an
extracted 10-gram will be highly recognizable as stemming from the
input. This nonlinearity is in contrast
to \newcite{weber2018controlling}, which used a linear penalty to
control the amount of copying in a pointer network.

In decoding, we search for the summary $\hat{\vec{y}}$ that maximizes
the product of the summarization model probability,
$p_{\text{M}}(\vec{y}\mid\vec{x})$, and the discount probabilities of
the extractive fragments $\fragments$:

\vspace{-6mm}\begin{equation} \label{eq:argmax}
\hat{\vec{y}} = \argmax_{\vec{y}} p_{\text{M}}(\vec{y}\mid\vec{x}) 
 \times \prod_{f\in\fragments} \lambda_h(|\vec{f}|)
\end{equation}

\begin{figure*}[t] 
  \centering
  \includegraphics[width=\textwidth]{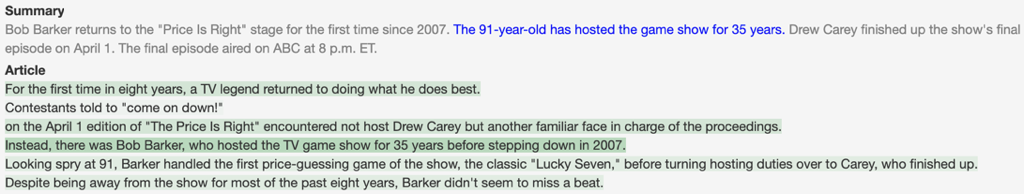}
  \caption{Screenshot (part) of a Mechanical Turk task (HIT) to judge
    the factuality of a summary sentence (in blue) with respect to
    news articles. Darker green article sentences are more similar to
    the blue summary sentence.  The full task showed sentences from
    two more articles in the same cluster; from the Multi-News test
    set.}
  \label{fig:mturk1}
\end{figure*}

\textbf{Beam Decoding.}
The model probability $p_{\text{M}}(\vec{x}, \vec{y})$ in neural text
generation models (\secref{exp-setup}) decomposes 
for token-by-token decoding as
$\prod_{i=1}^{|\vec{y}|}p_{\text{M}}(y_i\mid \vec{x}, y_1, \dots,
y_{i-1})$. Similarly, we decompose the application of the $\lambda_h$
function for any partial or completed extractive fragment $\vec{f}$:

\vspace{-3mm}\begin{equation} \label{eq:lambdadivide}
\lambda_h(|\vec{f}|) = \prod_{l=1}^{|\vec{f}|} \frac{\lambda_h(l)}{\lambda_h(l-1)}
\end{equation}

Therefore, to successively apply $\lambda_h$ at each output position
$i$ in beam decoding, each candidate for token $y_i$ is evaluated to
check whether choosing it would extend an extractive fragment to
length $l$. If so, its model probability $p_\text{M}(y_i\mid\dots)$ is
multiplied with $\lambda_h(l)$ and the $\lambda_h(l-1)$ that was
applied to the previous token $y_{i-1}$ is divided out. We are not
aiming to control the length of the generated output; instead we
penalize the model in proportion to the length of any phrases it would
extract from the input and encourage it to use novel phrases instead.

\noindent\textbf{Extraction Rewards.} We can choose to apply an extraction \textit{reward}, rather than a
penalty, by using the inverse $1/\lambda_h$; smaller values of $h$
then result in summaries that are more \textit{extractive}.

\section{Factuality}\label{sec:factuality}

We now describe metrics for factuality, before we can describe the
relationship between abstractiveness and factuality
(\secref{tradeoff}). By factuality of a summary $\vec{y}$, we mean
factual consistency with the input $\vec{x}$, rather than objective
factuality or universal truth. Measuring factuality automatically is
an active area of research \cite{Gabriel2020-gofigure}. Factuality is
most naturally measured by human annotators; we describe our setup for
human factuality annotation first, then move to automatic metrics.

\subsection{Human-annotated Factuality}\label{sec:mturk}

We use Amazon's Mechanical Turk (AMT) to measure the factuality of
automatically generated summaries with human annotators. These
annotators are untrained, so we use multiple mitigation strategies to
obtain high-quality judgements. We simplify the task: To avoid
overwhelming annotators with long text, we select a single sentence
per summary and ask the annotators if it is factually consistent with
the shown article(s). The other sentences of the summary are given as
well for context, shown in gray (see \xfigref{mturk1}).  The article(s)
are shortened to show a total of 9 sentences that were determined to
be semantically most similar to the selected summary
sentence;\footnote{We measure cosine similarity of sentence encodings
computed by the Universal Sentence Encoder \cite{Cer2018-universal}.}
the remaining article parts are replaced by ``\dots''. The summary
sentence is selected at random in proportion to its length. For each
summary, we get judgements only for the randomly selected
sentence. Aggregated over a set of summaries, we measure the average
chance of any randomly selected summary sentence to be factual. We
have verified high correlation of these factuality rates with the
factuality rates obtained through professional annotators who judged
complete summaries with respect to the full articles
(see \appref{mturk}).

We provide detailed task instructions, including examples for
intrinsic and extrinsic factual errors
\cite{maynez-etal-2020-faithfulness}. We require that potential
annotators pass a custom qualification test of finding factuality
errors.  Only workers with at least 100 completed tasks on AMT with an
acceptance rate of 95\%+ may take the test; 15\% of those pass,
enabling them to work on our tasks.  We use three annotators per task
and use MACE \cite{hovy-etal-2013-mace} to aggregate annotations and
recover the most likely binary factuality judgement per summary. We
add summaries for which we know the correct factuality annotation and
repeatedly check the annotators' accuracy on those summaries while
they are annotating; all answers from annotators who fall below a
threshold are replaced by answers from additional
annotators. Appendix~\ref{appendix:mturk} describes more details on
our setup and fair compensation.

For any set of generated summaries, we create the AMT tasks, get an
aggregate binary judgement per summary based on the multiple answers
as described, and report the mean of all human binary summary
factuality judgements; we call this score \textbf{\facth}
(\tabref{facth-f50}). We collect human factuality judgements for 10.2k
\bart summaries with varying degrees of abstractiveness, and for 4.2k
summaries from five different summarization models.

\textbf{Released Datasets.} We release
these human judgements as datasets called \textsc{ConstraintsFact}
(\secref{across-datasets}) and \textsc{ModelsFact}
(\secref{across-models}). Previous datasets with human factuality
judgements \cite{wang2020asking,kryscinski2020evaluating,maynez-etal-2020-faithfulness,pagnoni-etal-2021-understanding}
are substantially smaller, with under 5k summaries each, and
our \textsc{ConstraintsFact} dataset is the first that evaluates the
factuality of summaries with systematically varied degrees of
abstractiveness.

\subsection{Automatically Measured Factuality}\label{sec:factuality-metrics}

Measuring factuality \textit{automatically} is an active research
area; \newcite{pagnoni-etal-2021-understanding} gives an overview over
recent metrics and compares their correlations to human judgements,
where \textbf{DAE} \cite{goyal-durrett-2020-evaluating,goyal-durrett-2021-annotating}
and \textbf{FactCC} \cite{kryscinski2020evaluating} perform well. DAE
is an entailment model that classifies the factuality of the
dependency arcs in the summary, resulting in fine-grained judgements
at the subsentence level.  FactCC is a BERT-based binary classifier
trained on pairs of input and output sentences, where the output
sentence is annotated as either factual or non-factual.

\section{Abstractiveness-Factuality Tradeoff}\label{sec:tradeoff}

The metrics for factuality and abstractiveness along with the
abstractiveness constraints allow us to systematically explore the
relationship between abstractiveness and factuality. We can control
abstractiveness and observe the effect on factuality, i.e., we can
vary the amount of lexical overlap between input and generated summary
and observe the extent to which the summary preserves the input
semantics.

\begin{figure}[t] 
  \centering
  \includegraphics[width=\columnwidth]{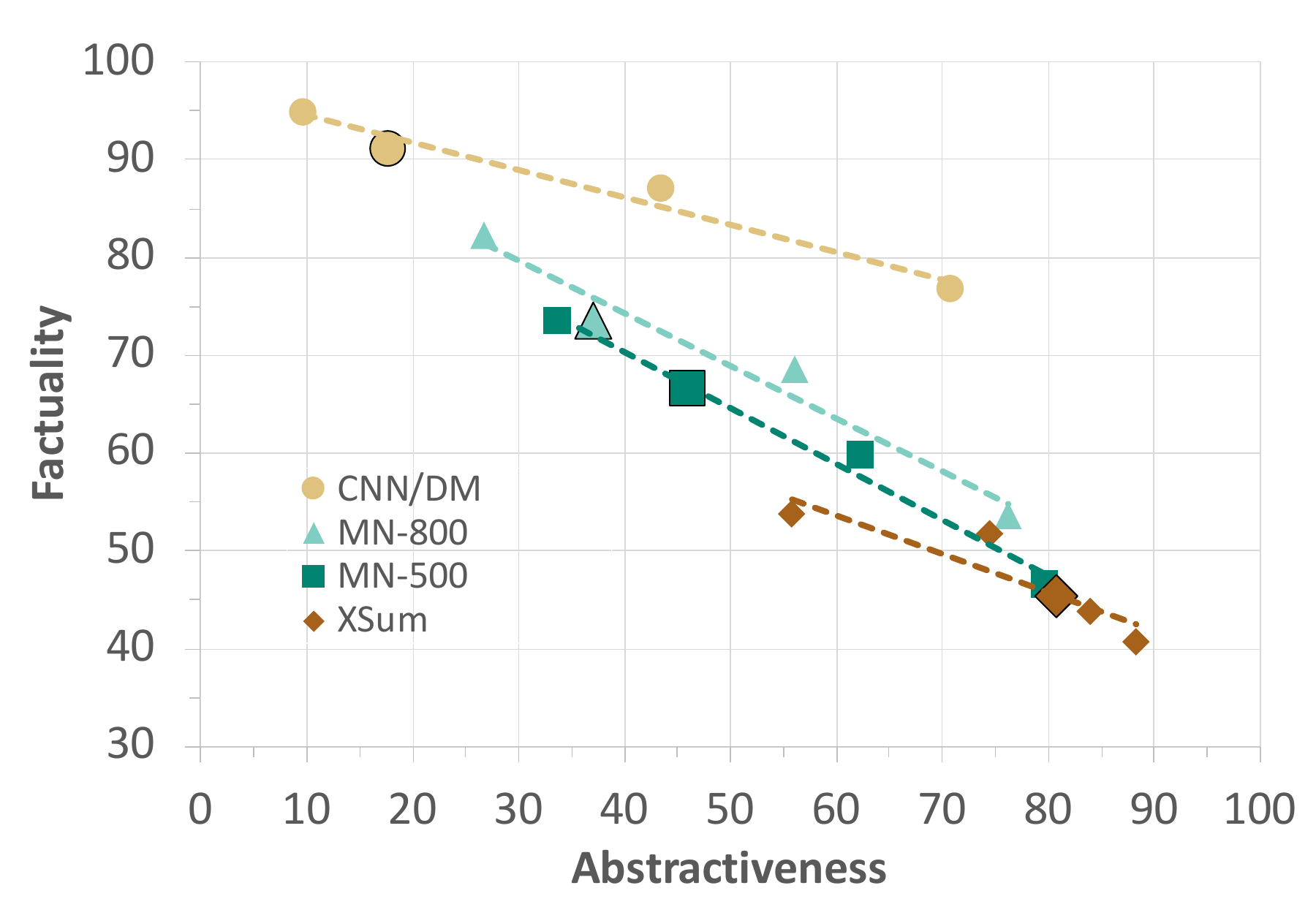}
  \caption{Human factuality judgements (\facth) for different degrees
    of abstractiveness (\mint).  Each color represents a \bart model
    trained on a particular dataset, decoded with varying decoding
    constraints (\sref{slider}); large outlined symbols mean no
    constraints.}
  \label{fig:trends}
\end{figure}

\paragraph{Factuality Trend Lines.}\label{sec:new-metrics}
To explore this relationship, we train summarization models on
different datasets. For any trained summarization model, we decode the
test set multiple times with different $h$ values for $\lambda_h$
(\eqref{lambda}),  resulting in sets of summaries with varying degrees
abstractiveness.  For each of these test set decodings, we measure
abstractiveness using \mint and the corresponding factuality using
human annotations, unless otherwise noted. This results in a series of
(abstractiveness, factuality) points for any trained summarization
model, which can be plotted, along with a linear trend
line. \xfigref{trends} shows such a plot; \secref{exp-results}
discusses its details.

\paragraph{F@50 Score.} Given each trend line, we can read off the factuality at 50\% abstractiveness, an
intuitively interpretable metric, which we call F@50; it provides a
comparison of the factuality of different models with a fixed degree
of abstractiveness.

\paragraph{\mint-adjusted Factuality Scores.} We characterize the tradeoff on
any single decoding output using a weighted average between factuality
and abstractiveness, $(\phi F + A)/(\phi+1)$. To measure
abstractiveness $A$, we use \mint; to measure factuality $F$, we use
human-measured factuality or an automatic metric with [0,1] range like
DAE or FactCC, resulting in abstractiveness-adjusted factuality
metrics \textbf{$\mu$FactH}, \textbf{$\mu$DAE}, \textbf{$\mu$FactCC}, etc.

We give factuality a higher weight, since factual semantic
representation of the input is a fundamental requirement for
summarization and low factuality can have negative societal
impact \cite{Zellers2019Defending}, while abstractiveness is a
desirable stylistic property.  When two measures are combined into one
comprehensive evaluation metric there is no \textit{a priori} correct
mixture weight; we follow common practice to give the more important
measure twice the
weight \cite{kohonen-etal-2010-semi,li-etal-2020-context-aware,preuss-etal-2021-automatically,opitz-frank-2021-towards}
and set $\phi$ to 2. By this definition, a system
whose factuality decreases by $x$ units, as compared to another
system, must make up for the lost factuality by $2x$ units in
abstractiveness to get the same score. When two systems have the same
factuality, the score prefers the one with higher abstractiveness.

\subsection{Discussion}

The abstractiveness-adjusted factuality metrics address the issue that
in the past, factuality rates of different systems have been compared
without taking abstractiveness into account. However, if one system
has a higher factuality rate than another, it may have achieved this
by copying phrases from the input into the summary with minimal
rephrasing, i.e., by having a low degree of abstractiveness. Such a
system may produce high-quality summaries, but their factuality rate
cannot directly be compared to the factuality numbers of more
abstractive summarization systems. Summarization methods that are
highly factual and abstractive are able to rephrase the input with few
factual errors; when we compare the factuality of abstractive
summarizers we must control for the amount of such rephrasing. The
abstractiveness-adjusted factuality metrics we propose enable us to
compare the factuality of abstractive summarization models even when
they perform different amounts of rephrasings.

As an analogy, consider precision and recall. High precision can be
trivially achieved with low recall, just as high factuality can be
achieved with low abstractiveness. Therefore when comparing the
precision of different retrieval systems, their recall numbers are
taken into account by using the F-score.\footnote{In our case, we use
a weighted arithmetic mean instead because an F score would steeply
decline to zero as abstractiveness goes to zero, which is undesirable
for output whose factuality is high.} Similarly, we argue that
factuality comparisons must take abstractiveness into account.

\section{Experiments}\label{sec:experiments}

\begin{table}[t]
\setlength{\tabcolsep}{8pt}
\centering\footnotesize
\ra{1}
\resizebox{\columnwidth}{!}{%
\begin{tabular}{@{}lrrrrr@{}}\toprule

  & $\lambda$ & \mint & \facth & \muFactH & F@50 \\
 \midrule

 \parbox[t]{1mm}{\multirow{4}{*}{\rotatebox[origin=c]{90}{CNN/DM}}}
 & $1/\lambda_2$ &       9.7 &    94.8 &      66.5 & \multirow{4}{*}{84.4}\\
 & none          &      17.6 &    91.2 &      66.7 \\
 & $\lambda_4$   &      43.5 &    87.0 &      72.5 \\
 & $\lambda_2$   &      70.8 &    76.7 &      74.7 \\
 
 \addlinespace[3mm]
 \parbox[t]{1mm}{\multirow{4}{*}{\rotatebox[origin=c]{90}{MN-800}}}
 & $1/\lambda_2$ &      26.8 &    82.2 &      63.7 & \multirow{4}{*}{68.9}\\
 & none          &      37.0 &    73.5 &      61.3 \\
 & $\lambda_4$   &      56.1 &    68.5 &      64.4 \\
 & $\lambda_2$   &      76.2 &    53.5 &      61.1 \\

 \addlinespace[3mm]
 \parbox[t]{1mm}{\multirow{4}{*}{\rotatebox[origin=c]{90}{MN-500}}}
 & $1/\lambda_2$ &      33.6 &    73.5 &      60.2 & \multirow{4}{*}{64.4}\\
 & none          &      45.9 &    66.5 &      59.6 \\
 & $\lambda_4$   &      62.3 &    59.7 &      60.6 \\
 & $\lambda_2$   &      79.7 &    46.5 &      57.6 \\
 
 \addlinespace[3mm]
 \parbox[t]{1mm}{\multirow{4}{*}{\rotatebox[origin=c]{90}{XSum}}}
 & $1/\lambda_1$ &      55.8 &    53.7 &      54.4 & \multirow{4}{*}{56.7}\\
 & $1/\lambda_2$ &      74.5 &    51.7 &      59.3 \\
 & none          &      80.8 &    45.3 &      57.2 \\
 & $\lambda_4$   &      84.0 &    43.7 &      57.1 \\
 & $\lambda_2$   &      88.3 &    40.7 &      56.5 \\
   
\bottomrule
\end{tabular}
}
\caption{Abstractiveness and factuality on 600 test samples per
  setting. The 17 \mint and \facth numbers are as shown in
  \xfigref{trends}; we add \muFactH and F@50.}\label{tab:facth-f50}
\end{table}

\subsection{Comparison Across Datasets Using NAC}\label{sec:across-datasets}

\begin{table}
\small\centering
\begin{tabular}{lccc}
\toprule
	Dataset & Train & Valid & Test \\ \midrule
	CNN/DM & 287,227 & 13,368 & 11,490 \\ 
	XSum & 204,045 & 11,332 & 11,334 \\ 
	Multi-News & 44,972 & 5,622 & 5,622 \\
\bottomrule
\end{tabular}
\caption{Train/valid/test split on public datasets.}\label{dataset_stats}
\end{table}

\textbf{Datasets.} We use CNN/DM \cite{hermann2015teaching},
	XSum \cite{narayan-etal-2018-dont}, and
	Multi-News \cite{fabbri-etal-2019-multi}, all of which contain
	English-only text. CNN/DM contains news articles from CNN and
	DailyMail paired with bullet point summaries.  XSum contains
	articles from BBC News, using each article's first sentence as
	summary.\footnote{Following \newcite{wang2020asking}, we
	reinsert the first sentences whenever we measure factuality of
	XSum summaries on AMT or with automatic metrics.} In
	Multi-News, each summary is written by a professional editor
	and paired with a cluster of news articles. For all three
	public datasets, we use the provided training/validation/test
	split.  The sizes of the three datasets are listed in
	Table~\ref{dataset_stats}.  From each of the three datasets,
	we use 600 samples to compare human and automatic factuality
	judgements.\footnote{For Multi-News and XSum, we take the
	first 600 samples per test set. For CNN/DM, we take the first
	300 and the last 300 test samples, from CNN and Daily Mail,
	respectively.}

\subsubsection{Setup}\label{sec:exp-setup}
We use the \bart \cite{lewis-etal-2020-bart} sequence-to-sequence model, 
which was pretrained on 160GB of text and gives competitive results on
CNN/DM and XSum. Our models use the provided model checkpoints for the
CNN/DM and the XSum datasets as well as the recommended decoding
settings.
For Multi-News (MN), we train a model on the training set, starting
from the \texttt{bart.large} pretrained model.\footnote{We train for
five epochs (learning rate: 2e-5) and limit output to 50 to 300
tokens.} For Multi-News, we truncate the input documents per cluster
so that their combined length does not exceed N words, following
\newcite{fabbri-etal-2019-multi}. We train models with $N=800$ and
$N=500$, called MN-800 and MN-500, respectively. We measure the \mint
scores for the reference summaries in these datasets; these can be
compared to the \mint scores obtained in decoding
(\secref{exp-results}). The test set references for MN-500 have a
\mint score of 78.2\%, compared to 72.8\% for MN-800.  \mint is higher
for MN-500 since the shorter truncation removes article content that
could otherwise overlap with the summaries. The \mint scores for the
CNN/DM and XSum references are 59.6\% and 87.8\%, respectively; XSum
is the most abstractive dataset.

\subsubsection{Results}\label{sec:exp-results}

We use each of the four BART models to decode its respective test set
multiple times, with varying abstractiveness constraints, resulting in
17 outputs. For each one, we obtain human factuality judgements on the
corresponding 600 samples, resulting in 17 x 600 human factuality
judgements -- our \textsc{ConstraintsFact} dataset --, which we
aggregate into 17 mean \facth scores; we also compute the
corresponding 17 \mint scores. \xfigref{trends} plots
the resulting abstractiveness and human-measured factuality for each
of the four models, thereby providing a visual representation of the
abstractiveness-factuality tradeoff for these
models. \tabref{facth-f50} shows the same 17 \mint and \facth values,
along with \muFactH and F@50 scores.

The lower right of \xfigref{trends} shows five lozenges
(\textcolor{xsumcolor}{$\blacklozenge$}). The larger one represents
the decoding with our \textbf{XSum}-trained model using default
settings; the other four red points represent decodings under the same
model, but with different abstractiveness constraints that result in
more \textit{extractive} ($1/\lambda_h$) or more \textit{abstractive}
($\lambda_h$) summaries (\secref{slider}). The five red points are
associated with a dashed linear trend line. Compared to the other
points in the figure, abstractiveness is high and factuality low --
the model tends to paraphrase its input, often incorrectly. It took a
strong extractive reward ($1/\lambda_1$), which we did not use for the
models trained on other datasets, to bias this model toward lower
abstractiveness and higher factuality.

For the \textbf{Multi-News} models, four decodings using MN-500 are
shown as squares (\textcolor{mn500color}{$\blacksquare$}), decodings
under MN-800 as triangles ({\LARGE
  \textcolor{mn800color}{$\blacktriangle$}}). The MN-800 model is more
factual across the abstractiveness spectrum. This can be explained by
the fact that for MN-500, larger parts of the input are truncated
(\secref{exp-setup}) that the untruncated reference summary in
training may still refer to; the MN-500 model learns to hallucinate
more.

The four decodings for \textbf{CNN/DM} are shown as bullets
(\textcolor{cnndmcolor}{\huge $\bullet$}). Its model output without
abstractiveness constraint (large bullet) is the most extractive; the
extraction reward to its left (using $1/\lambda_2$) cannot make it
much more extractive; however, there is room to the right, and the
abstraction rewards ($\lambda_4$ and $\lambda_2$) move its
abstractiveness far into the abstractiveness level of Multi-News and
XSum.

\paragraph{F@50 Scores.}
One of the main takeaways of this study is that different systems can
have different factuality rates at the same level of
abstractiveness. Previous authors have observed that XSum summaries
are highly abstractive and less factual, and that CNN/DM summaries are
at the opposite side of that spectrum. We confirm this; however, we
add that we can bias the XSum model to create less abstractive
summaries and the CNN/DM model to create more abstractive models, so
that \textbf{their abstractiveness becomes comparable}, and the
factuality rates still differ considerably: Based on the trend line,
the F@50 score of the XSum model is 56.7\%, while the CNN/DM model's
F@50 is 84.4\%. MN-800 and MN-500 lie in the middle.

\paragraph{\muFactH Scores.}
The \muFactH scores adjust \facth for abstractiveness. They penalize
the CNN/DM model for its low abstractiveness and reward the XSum model
for its high abstractiveness, bringing them closer together, compared
to their more divergent \facth scores. The \muFactH scores for MN-800
and MN-500 are also close (59.6\% versus 61.3\% for $\lambda$=none),
as MN-800 is more factual but also less abstractive.

\begin{table}[t]
\setlength{\tabcolsep}{4pt}
\centering\footnotesize
\ra{1}
\begin{tabular}{@{}lr@{\hskip 7mm}r@{\hskip 5mm}rrrrr}\toprule
  & $\lambda$ & RL & \mint & p3 & p4 &  lcsr &   density \\
  
  \midrule

  \parbox[t]{1mm}{\multirow{4}{*}{\rotatebox[origin=c]{90}{CNN/DM}}}
  & $1/\lambda_2$ & 37.9 &    9.0 & 89.0 & 84.7 &   93.1 &      28.9 \\
  & none          & 41.0 &   16.8 & 79.5 & 72.1 &   89.4 &      15.4 \\
  & $\lambda_4$   & 41.5 &   43.7 & 50.0 & 35.1 &   77.8 &       4.6 \\
  & $\lambda_2$   & 39.3 &   70.3 & 26.4 & 12.6 &   67.4 &       2.2 \\
  \addlinespace[3mm]

  \parbox[t]{1mm}{\multirow{4}{*}{\rotatebox[origin=c]{90}{MN-800}}}
  & $1/\lambda_2$ & 44.8 &   26.6 & 71.1 & 64.1 &   69.5 &      20.7 \\
  & none          & 45.8 &   37.1 & 58.9 & 50.1 &   63.3 &      13.4 \\
  & $\lambda_4$   & 45.8 &   56.3 & 38.7 & 27.0 &   51.9 &       4.3 \\
  & $\lambda_2$   & 44.0 &   76.4 & 20.7 & 10.4 &   41.6 &       2.0 \\
  \addlinespace[3mm]

  \parbox[t]{1mm}{\multirow{4}{*}{\rotatebox[origin=c]{90}{MN-500}}}
  & $1/\lambda_2$ & 44.6 &   34.1 & 63.7 & 56.4 &   61.0 &      17.6 \\
  & none          & 45.5 &   45.9 & 50.2 & 41.4 &   54.2 &      10.6 \\
  & $\lambda_4$   & 45.1 &   62.2 & 33.4 & 22.7 &   44.8 &       3.6 \\
  & $\lambda_2$   & 43.3 &   79.8 & 17.8 &  8.8 &   35.9 &       1.8 \\
  \addlinespace[3mm]

  \parbox[t]{1mm}{\multirow{4}{*}{\rotatebox[origin=c]{90}{XSum}}}
  & $1/\lambda_1$ & 30.8 &   53.8 & 41.7 & 32.3 &   66.9 &       5.8 \\
  & $1/\lambda_2$ & 36.0 &   73.9 & 23.0 & 14.1 &   57.7 &       3.0 \\
  & none          & 36.8 &   80.2 & 17.6 &  9.2 &   54.5 &       2.4 \\
  & $\lambda_4$   & 36.8 &   83.6 & 14.6 &  6.6 &   52.8 &       2.2 \\
  & $\lambda_2$   & 36.3 &   88.1 & 10.8 &  4.1 &   49.8 &       1.9 \\

  \bottomrule
\end{tabular}
\caption{Impact of $\lambda$ on \rouge-L F$_1$ (RL) and abstractiveness metrics on the full test sets. p3, p4, lcsr are component scores in \mint (\sref{mintscore}), density is average length of extracted fragments \cite{grusky-etal-2018-newsroom}. \rouge measures overlap with reference summaries, abstractiveness metrics measure input overlap.}\label{tab:rouge-mint}
\end{table}

\paragraph{Summary Quality and Abstractiveness.}
\tabref{rouge-mint} lists \rouge-L scores for the different decodings,
along with abstractiveness metrics, measured on the \textit{full} test
sets. \rouge scores aim to measure summary quality by comparing the
generated summaries with the reference summaries, while
abstractiveness metrics measure overlap between the generated
summaries and the input.  Decodings without abstractiveness
constraints replicate previous works' \rouge scores
\cite{lewis-etal-2020-bart,fabbri-etal-2019-multi} (\appref{rouge}).  The $\lambda_4$ constraint can \textbf{dramatically
  increase abstractiveness while leaving \rouge scores virtually
  unchanged}.  We also conduct a human evaluation of informativeness
  and coherence, comparing unconstrained summaries with summaries
  generated with the $\lambda_4$ decoding constraint; the
  unconstrained decoding is preferred for XSum but the constrained
  decoding is preferred for CNN/DM, and results are mixed for
  Multi-News, see
\appref{baselines-quality}.  The density scores
\cite{grusky-etal-2018-newsroom} in the table have high correlation
with the \mint scores.

\subsection{Comparison Across Different Models}\label{sec:across-models}

\begin{table}[t]
\setlength{\tabcolsep}{5pt}
\centering
\ra{1}
\resizebox{\columnwidth}{!}{%
    \begin{tabular}{@{}llrr@{\hskip 2mm}cc@{}}\toprule
  
  & Model
  & \mint
  & \textcolor{mintgreen}{$\mu$}\facth
  & \textcolor{mintgreen}{$\mu$}DAE
  & \textcolor{mintgreen}{$\mu$}FactCC
  \\\midrule

  \parbox[t]{1mm}{\multirow{5}{*}{\rotatebox[origin=l]{90}{ \small CNN/DM}}}

  &  \textsc{Bart}     & 16.8     & \mucell{\bf 66.4}{91.2} & \mucell{\bf 67.4}{ 92.6} & \mucell{56.2}{ 75.9}     \\
  & \textsc{BertSum}  & 14.1     & \mucell{64.7}{90.0}     & \mucell{57.8}{79.6}     & \mucell{57.0}{78.5}     \\
  & \textsc{PGConv}   & 5.5      & \mucell{63.5}{\bf 92.5} & \mucell{64.0}{\bf 93.3} & \mucell{62.3}{\bf 90.7} \\
  & \textsc{BottomUp} & 17.2     & \mucell{50.6}{67.3}     & \mucell{55.0}{73.9}     & \mucell{54.3}{72.9}     \\
  & \textsc{AbsRL}    & \bf 18.9 & \mucell{60.6}{81.5}     & \mucell{62.3}{84.0}     & \mucell{\bf 64.1}{86.8} \\
            
  \addlinespace[3mm]

  \parbox[t]{1mm}{\multirow{2}{*}{\rotatebox[origin=l]{90}{\small XSum}}}

   & \textsc{Bart}    & 80.2     & \mucell{\bf 56.9}{\bf 45.3} & \mucell{\bf 67.3}{\bf 60.8} & \mucell{\bf 53.9}{\bf 40.8} \\
   & \textsc{BertSum} & \bf 82.8 & \mucell{52.1}{36.8}         & \mucell{61.5}{50.8}         & \mucell{50.8}{34.8}         \\
      
  \bottomrule

\end{tabular}
}
\caption{Abstractiveness (\mint) and factuality of different
   models. For each factuality metric, we first list
  its \mint-adjusted variant in green. Example: \bart's \muFactH is
  66.4, while the unadjusted \facth is 91.2. All numbers are
  percentage scores $\in$ [0,100].}
\label{tab:comparing-systems-short}
\end{table}

We also compare the abstractiveness-factuality tradeoffs of
summarization models from the literature.  We obtain outputs of four
summarization models other than \textbf{\bart}:
\textbf{\textsc{BertSum}} \cite{Liu2019BertSum} is a transformer model
in which only the encoder is pretrained; \textbf{\textsc{PGConv}}
\cite{see2017get} is a pointer-generator network; 
\textbf{\textsc{BottomUp}} \cite{Gehrmann2018bottomup} and
\textbf{\textsc{AbsRL}} \cite{Chen2018AbsRL} select 
source fragments to constrain an abstractive generation model. We
obtain human factuality judgements of the five model outputs on 600
samples of CNN/DM and XSum, respectively, and release this as our
\textsc{ModelsFact} dataset; we apply automatic metrics (e.g., DAE) as well as
our abstractiveness-adjusted variants (e.g., $\mu$DAE) to the
\textit{full} test sets. \tabref{comparing-systems-short} shows the
results. For CNN/DM, we find that the highly extractive
model \textsc{PGConv} receives the highest automatic and human
factuality scores, while the abstractiveness-adjusted variants
favor \textsc{Bart} or \textsc{AbsRL}, whose outputs represent better
tradeoffs between abstractiveness and
factuality. On \textbf{XSum}, \bart's output is considerably more
factual than \textsc{BertSum}'s across all factuality metrics,
while \bart has only slightly lower abstractiveness; as a
result, \bart is also favored by all \mint-adjusted factuality
metrics. Detailed results including additional factuality metrics are
described in \appref{across-models}.

The \mint-adjusted variants of factuality metrics put factuality rates
into perspective. We encourage authors who compare factuality rates
across summarization models to also compare \mint-adjusted variants
(e.g., $\mu$DAE), to account for differing levels of abstractiveness.

\section{Related Work}

\textbf{Abstractiveness-Factuality Tradeoff:} \newcite{durmus2020feqa}
observe that abstractiveness at test time depends on the
abstractiveness of the training data and that highly abstractive
summaries tend to be less factual. We control for abstractiveness and
see that factuality rates between different systems can vary widely at
the \textit{same} abstractiveness
levels. Recently, \newcite{ladhak2022faithful} present an alternative
framework to evaluate the faithfulness-extractiveness tradeoff,
requiring training multiple models on subsets of the training data to
measure the tradeoff, while we use constraints to analyze tradeoffs
that a single model makes. \textbf{Increasing Abstractiveness:}
\newcite{kryscinski-etal-2018-improving} use policy gradient with a
novelty reward to encourage abstraction in a pointer-generator (PG)
\cite{gulcehre2016pointing, see2017get}.
\newcite{weber2018controlling} penalize copying tokens
during PG decoding.  Our constraints apply to general
sequence-to-sequence models and include nonlinear
penalties. \newcite{song2020controlling} control copying in training
abstractive summarization models by masking the summary tokens with
different probabilities, depending on whether they are seen in the
input document or not. In contrast, our technique does not require
retraining to obtain varying degrees of abstractiveness.

\section{Conclusions}

We presented new metrics and datasets for evaluating the relationship
of abstractiveness and factuality. As part of our analysis, we
presented abstractiveness constraints, which can bias a summarization
model to increase or decrease the level of abstractiveness while
generating summaries, using nonlinear penalties or rewards based on
the length of summary fragments extracted from the source. Through
automatic and human factuality evaluations, including 10.2k human
factuality judgements of summaries with systematically varied
abstractiveness, we shed light on how abstractiveness interacts with
factuality, across multiple datasets and models.  We proposed new
metrics to measure the tradeoff, including F@50 and \mint-adjusted
factuality rates, such as $\mu$DAE and $\mu$FactCC, and we established
baselines for future research.

\section*{Limitations}

The abstractiveness constraints we have presented can be used to
increase or decrease the abstractiveness of the generated
text. Dedicated code is needed to integrate such constraints into a
decoder. The constraints are needed to obtain trend lines as
in \xfigref{trends}, as well as the F@50 score. However,
the \mint-adjusted factuality scores, such as $\mu$FactH, $\mu$DAE or
$\mu$FactCC can be computed for any summarization system, without the
need for implementing abstractiveness constraints, as we have done
in \secref{across-models}.

\section*{Ethical Considerations}

We have analyzed the factuality of generated text in relation to the
abstractiveness of the source texts; we have also proposed new metrics
that let researchers compare the factuality of different generative
models. As such, we consider our work a contribution toward text
generation methods that make fewer factual mistakes and become
therefore more reliable and responsible. However, any advance in text
generation methods can be used by bad actors to cheaply generate
misleading or harmful texts.

We hired annotators on the Mechanical Turk platform to judge
machine-generated summaries. Our first ethical consideration with
respect to this data collection is fair and prompt pay for the work of
the annotators. We describe in Appendix C that we paid all human
subjects a fair average pay of \$12.50 USD per hour, based on observed
median time spent per HIT. As described (Section 3.1), we
automatically approved the annotators' work promptly and paid bonuses
as appropriate. The annotators' privacy and confidentiality were
respected at all times.

\bibliographystyle{acl_natbib}
\interlinepenalty=10000
\bibliography{factual_abstractive,mengwen}

\clearpage

\appendix

\section{Measuring Abstractiveness with \mint}\label{appendix:mint}

\paragraph{$N$-gram Overlap.}
Each $p_n$, short for $p_n(\vec{x}, \vec{y})$, is the \ngram precision
of the \ngrams in $\vec{y}$ with respect to $\vec{x}$, i.e., the
percentage of \ngrams in $\vec{y}$ that are extracted from
$\vec{x}$.\footnote{\mint has elements of \rouge \cite{lin-2004-rouge}
  and \bleu \cite{papineni-etal-2002-bleu}.  We do not use the
  \textit{modified} \ngram precisions, like \bleu does, because
  \ngrams extracted multiple times from $x$ should count as such every
  time.} For highly abstractive outputs, higher-order \ngram precision
can be zero, leading to an undefined or zero harmonic mean value.  We
prevent this by smoothing the \ngram counts from which \ngram
precisions are calculated, such that each \ngram count is the average
of itself and the smoothed ($n-1$)-gram count and the unsmoothed
($n+1$)-gram count. The smoothed $0$-gram count is defined as
  the $1$-gram count plus one.  We chose this method for its
simplicity and effectiveness; it is described as method 5 in
\newcite{chen-cherry-2014-bleu}. 

\paragraph{Harmonic Mean.}
We use the harmonic mean, in analogy to the definition of the F$_1$
score, as it is a mean function designed to aggregate ratios with
different denominators.

For a completely extractive summary that extracts sentences in the
original order, the \mint score is $0$. The score increases as the
order of the extractive fragments is changed with respect to the
input, their lengths are decreased and new words and fragments are
introduced that are not part of the input $\vec{x}$.  The use of the
length-normalized LCS score (lcsr) is inspired by \rouge-L; it is a
useful addition to the \ngram precisions as it can detect the
extraction of longer \ngrams broken up by minor edits. As an example,
consider the $(\vec{x}, \vec{y})$ pair shown in
\xfigref{fragments-example}.  Only 4 of the 12 summary fourgrams match
the input, i.e., $p_4$=33.3\%, although very high overlap is apparent
due to the fact that a 15-word fragment from the input was extracted
with only the words ``verdict'' and ``which'' minimally changed by
synonym substitution. The lcsr score reflects this and measures
12/15=80.0\% overlap. On the other hand, the \ngram precisions used in
the \mint score are valuable in detecting textual overlaps that are
not part of the longest common subsequence.

\section{Details on the Abstractiveness
Constraints}\label{appendix:constraints}

\paragraph{Log Space.}
We have described the abstractiveness constraints in probability
space.  In practice, we equivalently search for $\hat{\vec{y}}$ in log
space using log probabilities and the log of $\lambda_h$ defined in
\eqref{lambda}. It can be shown that
$\log\lambda_h(|\vec{f}|)=\frac{-|\vec{f}|^2}{(1.20112\times h)^2}$.

\section{Details on Our Mechanical Turk Setup}\label{appendix:mturk}

\begin{figure*}[t] 
  \centering
  \includegraphics[width=\textwidth]{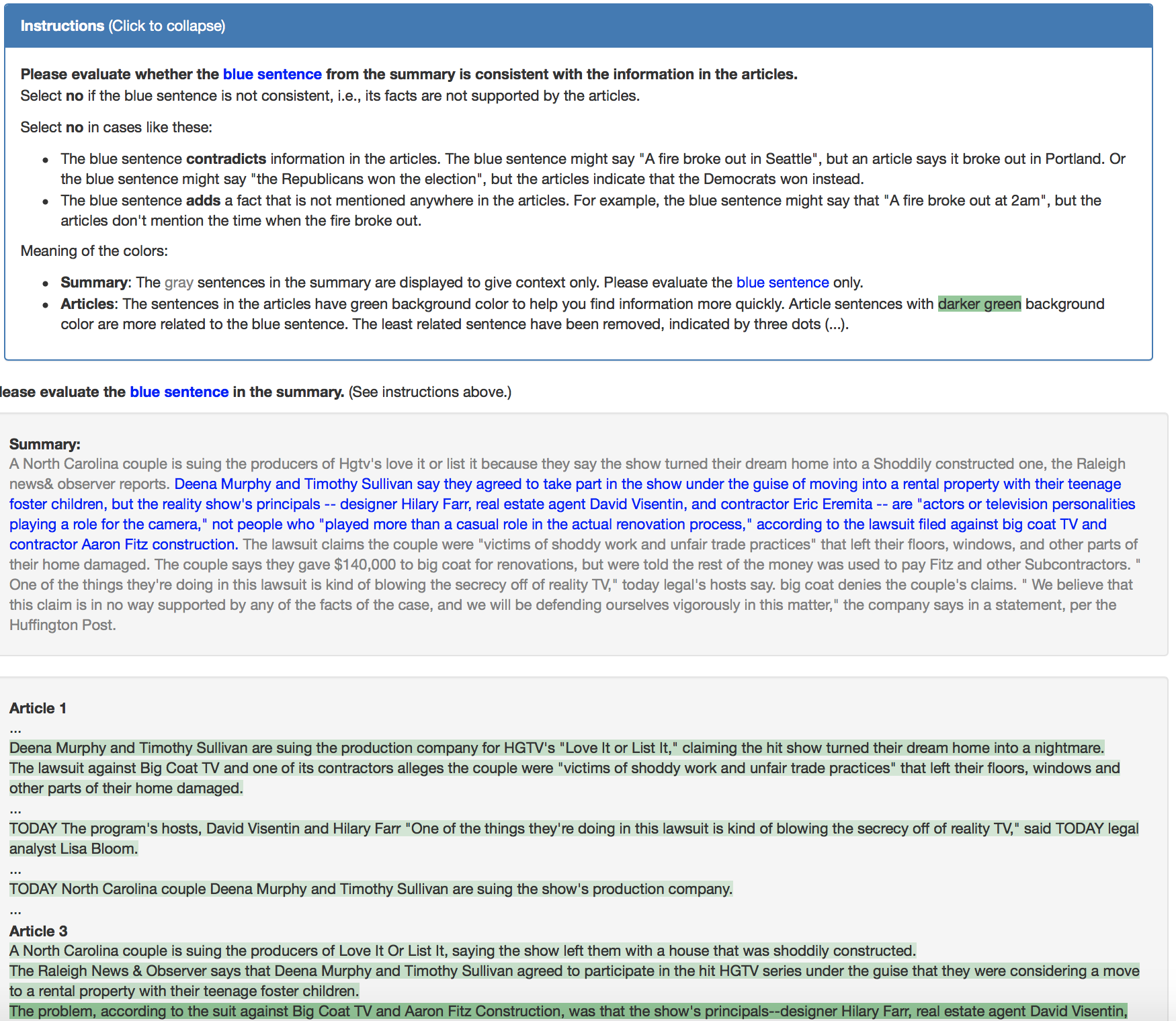}
  \caption{Instructions for the factuality annotation task on Amazon
    Mechanical Turk, as well as the summary and part of the article
    text shown to the worker.}
  \label{fig:mturk-factual-instructions}
\end{figure*}

We provide additional details on the strategies we use to obtain
high-quality judgements on Amazon Mechanical Turk. We give detailed
instructions to the annotators, with definitions and examples of
different factual errors (see \xfigref{mturk-factual-instructions}). We
also add a request to write a short explanation when a sentence is
judged as not factual.

\paragraph{Tasks with Known Answers.}
We add a number of tasks with known answers, enabling us to estimate
the accuracy of workers who work on multiple of these. 

\paragraph{Automatic Quality Checks.}
Workers who complete the tasks too quickly, write no or very short
explanation texts or have low accuracy on the tasks with known answers
are automatically removed from our worker pool. Their answers are
replaced with new answers.

\paragraph{Bonus.}
We use a bonus incentive structure. Every worker who passes the
automatic quality checks receives a bonus at the end.

\paragraph{Check Against Professional Annotators.}
We have seven sets of 150 automatically generated summaries each,
which we had previously sent to professional news editors to annotate
factuality. Those annotators rated the complete summaries with respect
to the complete inputs -- no sentences were preselected to simplify
the task. We re-annotated these summary-article pairs using our
Mechanical Turk setup, and the resulting per-set factuality rates
correlated highly (r=.88) with those previously obtained from the
professional annotators (p$<.05$).

As a further quality check, we sent one set of 600 summaries to
Mechanical Turk twice, several weeks apart. The two factuality rates
obtained for that same set were close -- 91.2\% and 92.0\%.

\paragraph{Qualification Test.}
For all our evaluations on Mechanical Turk (see \secref{mturk}), we
first set up a short qualification test that can be taken by any
worker from a country whose main language is English, who has
completed 100 or more HITs so far with an acceptance rate of 95\% or
higher. The qualification test consists of just three questions from
our factual consistency setup; two of which must be answered
correctly, along with an explanation text (5 words or more) to explain
when ``not factually consistent'' was chosen. 53\% of workers who
start the test provide answers to all three questions, and 27.6\% of
these answer at least two correctly and provide a reasonable
explanation text, i.e., only 14.6\% of the test takers are granted the
qualification.

The qualification enables workers to work on our factual consistency
HITs as well as our HITs judging informativeness and coherence.

\paragraph{Fair Compensation.}
The factual consistency task pays \$0.15 per HIT with a bonus
of \$0.05. It can be done quickly, given the fact that a single
summary sentence is evaluated and the related sentences in the article
are highlighted.  The task of evaluating informativeness and coherence
(see \appref{baselines-quality}) pays \$0.50 per HIT with a bonus
of \$0.25, as more text is displayed, compared to the factuality task.
These amount to an average pay of \$12.50 per hour, including the
bonus, based on median time spent per HIT. The bonus is paid to
workers who spend at least 10 seconds per HIT, give short explanation
texts for their decisions and maintain high accuracy on HITs with
known answers.

\section{Human Evaluation of Informativeness and Coherence}\label{appendix:baselines-quality}

We conduct a human evaluation to determine the informativeness and
coherence of the summaries generated with the $\lambda_4$ decoding
constraint (\eqref{lambda}), which increases abstractiveness, as
compared to not using any abstractiveness constraint. We use the same
setup as for the factuality task, including a qualification test,
three annotators per task and aggregation using MACE.

We use the following definitions of \textit{informativeness} and
\textit{coherence} for the human evaluation:
\begin{itemize}
    \item \textit{Informativeness}: The more informative summary is
      better at expressing the main points of the news story. It
      contains information that is more relevant and important. It has
      fewer unimportant details. Its content is more similar to the
      human-written summary.
    \item \textit{Coherence}: The more coherent summary has better
      structure and flow, is easier to follow. The facts are presented
      in a more logical order.
\end{itemize}

The results are shown in \tabref{baselines-quality}. For the
\textbf{CNN/DM} model, the output without decoding constraints is the
most extractive, and the raters preferred the more abstractive version
generated with the decoding constraint, both for informativeness and
coherence. For the \textbf{XSum} model, where the output with the
decoding constraint disabled is already highly abstractive, the result
is reversed. For \textbf{Multi-News}, the result is mixed: Raters
found the output with no decoding constraints more informative, but
less coherent.

\begin{table}
\centering\small
\begin{tabular}{lrrrrrr}
\toprule
                    & \multicolumn{2}{c}{CNN/DM} & \multicolumn{2}{c}{MN-800}& \multicolumn{2}{c}{XSum} \\
                    & inf. & coh.&  inf. & coh. &   inf. & coh. \\\midrule
prefer off         & 36.5 & 36.7 & 39.8 & 35.8 & 18.8 & 18.7 \\
prefer $\lambda_4$ & 46.5 & 39.2 & 34.7 & 39.8 & 16.5 & 16.3 \\
both equal         & 17.0 & 24.2 & 25.5 & 24.3 & 64.7 & 65.0 \\
\bottomrule
\end{tabular}
\caption{Human quality evaluation of summaries generated with no
  abstractiveness constraint (``off'') versus $\lambda_4$. We asked which
  summary is more informative or coherent, respectively. MN-800 stands
  for Multi-News with the input documents truncated to 800 words total
  (\secref{exp-setup}).}\label{tab:baselines-quality}
\end{table}

\section{More On Automatic Factuality Metrics}\label{appendix:more-automatic-metrics}

When we apply FactCC to a summary, we apply it separately to each
summary sentence and use the mean score per summary. For each sentence
that we score with FactCC, we shorten the input document by selecting
ten sentences with the highest cosine embedding
similarity \cite{conneau-etal-2017-supervised}, in order to fit the
input to the length limits.

In the following two appendix sections, we use not only DAE and
FactCC, as described in the main text, but also two metrics based on
question answering: FEQA \cite{durmus2020feqa} and
QAGS \cite{wang2020asking}.
\textbf{FEQA} generates questions from masked summary sentences whose masked entities are used as ``gold'' answers; these are compared to the answers obtained from a QA model on the input.
In \textbf{QAGS}, a question generation model generates questions from
the summary, a QA model answers these questions from both summary and
input, and the similarity of the answer pairs is evaluated.

\section{Correlating Human and Automatic Factuality Judgements}\label{appendix:correlations}

\tabref{model-correlations} shows correlations of the human judgements with different automatic metrics on the \textsc{ModelsFact} dataset, complementing earlier studies
\cite{Gabriel2020-gofigure,pagnoni-etal-2021-understanding}.
We compute correlations at the level of individual summaries. To make
meaningful comparisons between the human and the automatic scores, we
apply the automatic metrics here to the \textit{single} randomly
selected sentence per summary that the human annotators judged.
Overall, we observe here that DAE has the highest correlations with
human judgements.

\begin{table}[]
\setlength{\tabcolsep}{6pt}
\centering\small
\ra{1}
  \begin{tabular}{@{}lccccccc@{}}
    \toprule
    Data & Size & DAE & FactCC &  FEQA &  QAGS &
     \\ \midrule

    All    &  4.2k &  .44 & .35 &           .27 &   .44      \\
    CNN/DM &  3.0k &  .35 & .24 &           .05 &   .27      \\
    XSum   &  1.2k &  .39 & .17 &          $\dagger$.01 &   .25  \\

    \bottomrule \end{tabular} \caption{Pearson correlations to human
  factuality judgements on the \textsc{ModelsFact} dataset. The result
  with the $\dagger$ symbol is not
  significant.}  \label{tab:model-correlations} \end{table}

\section{Comparison Across Different Models}\label{appendix:across-models}

\begin{table*}[t]
\setlength{\tabcolsep}{5pt}
\centering
\ra{1}
\begin{tabular}{@{}clrr@{\hskip 5mm}ccccc@{}}\toprule
  Data
  & Model
  & \mint
  & \textcolor{mintgreen}{$\mu$}\facth
  & \textcolor{mintgreen}{$\mu$}DAE
  & \textcolor{mintgreen}{$\mu$}FactCC
  & \textcolor{mintgreen}{$\mu$}FEQA
  & \textcolor{mintgreen}{$\mu$}QAGS
  \\\midrule

  \parbox[t]{1mm}{\multirow{5}{*}{\rotatebox[origin=l]{90}{\small CNN/DM}}}

  & \textsc{Bart}     & 16.8     & \mucell{\bf 66.4}{91.2} & \mucell{\bf 67.4}{92.6} & \mucell{56.2}{75.9}              & \mucell{47.2}{62.4}     & \mucell{\bf 61.7}{84.2} \\
  & \textsc{BertSum}  & 14.1     & \mucell{64.7}{90.0}     & \mucell{57.8}{79.6}     & \mucell{57.0}{78.5}              & \mucell{47.6}{64.4}     & \mucell{60.8}{84.2}     \\
  & \textsc{PGConv}   & 5.5      & \mucell{63.5}{\bf 92.5} & \mucell{64.0}{\bf 93.3} & \mucell{62.3}{\bf 90.7}          & \mucell{45.2}{\bf 65.0} & \mucell{58.1}{\bf 84.4} \\
  & \textsc{BottomUp} & 17.2     & \mucell{50.6}{67.3}     & \mucell{55.0}{73.9}     & \mucell{54.3}{72.9}              & \mucell{47.3}{62.3}     & \mucell{58.2}{78.7}     \\
  & \textsc{AbsRL}    & \bf 18.9 & \mucell{60.6}{81.5}     & \mucell{62.3}{84.0}     & \mucell{\bf 64.1}{86.8}          & \mucell{\bf 49.6}{65.0} & \mucell{61.3}{82.5}     \\
            
  \addlinespace[3mm]

  \parbox[t]{1mm}{\multirow{2}{*}{\rotatebox[origin=l]{90}{\small XSum}}}

   & \textsc{Bart}    & 80.2     & \mucell{\bf 56.9}{\bf 45.3} & \mucell{\bf 67.3}{\bf 60.8} & \mucell{\bf 53.9}{\bf 40.8} & \mucell{\bf 50.9}{\bf 36.2} & \mucell{\bf 53.4}{\bf 40.1} \\
   & \textsc{BertSum} & \bf 82.8 & \mucell{52.1}{36.8}         & \mucell{61.5}{50.8}         & \mucell{50.8}{34.8}         & \mucell{46.6}{28.4}         & \mucell{46.0}{27.6}         \\
      
  \bottomrule

\end{tabular}
\caption{Abstractiveness (\mint) and factuality of different
  summarization models. For each factuality metric, we first list
  its \mint-adjusted variant in green. Example: \bart's \muFactH is
  66.4, while the unadjusted \facth is 91.2. All numbers are
  percentage scores $\in$ [0,100].}
\label{tab:comparing-systems}
\end{table*}

Here we offer an extended description of our comparison of the
abstractiveness-factuality tradeoffs of summarization models from the
literature, including the use of additional automatic factuality
metrics (see \appref{more-automatic-metrics}).

\tabref{comparing-systems} shows human and automatic factuality scores, as well as \mint-adjusted versions of these scores. 
We observe that all factuality metrics favor the output of
the \textsc{PGConv} model on \textbf{CNN/DM}; however, its low
abstractiveness indicates that its output falls into the ``trivially
factual'' quadrant (\xfigref{four-corners}). The \mint-adjusted
variants (shown in green) penalize such low abstractiveness, favoring
the \bart or \textsc{AbsRL} models instead, whose outputs represent
better tradeoffs between abstractiveness and factuality. Human
factuality raters (\facth) rank \textsc{AbsRL} in fourth place, while
FactCC, FEQA and QAGS rank it highly; we hypothesize
that \textsc{AbsRL} makes factual errors that these measures cannot
detect well.  On \textbf{XSum}, \bart's output is considerably more
factual than \textsc{BertSum}'s across all factuality metrics,
while \bart has only slightly lower abstractiveness; as a
result, \bart is also favored by all \mint-adjusted factuality
metrics. \bart's pretraining of both encoder and decoder may be
contributing to its factuality, in accordance with
\newcite{maynez-etal-2020-faithfulness}.
Note that for DAE, we apply the Ent-C model on CNN/DM output and the \textsc{XSum-Human} model on XSum output.
\appref{models-rouge} shows \textbf{\rouge} scores.

\section{\rouge Scores}\label{appendix:rouge}

\subsection{\bart Models}

The aim of this paper is not to improve \rouge scores, but to gain
insights about the tradeoff between abstractiveness and factuality. We
do, however, stress that the \bart models we use in our analysis are
competitive with the start of the art. We list our \rouge-1, \rouge-2
and \rouge-L F$_1$ scores, as well as their averages; see the RL
scores in \tabref{rouge-mint} as well: \itemsep0mm
\begin{itemize}
  \item For CNN/DM, our $\lambda$=none decoding has 44.1/21.2/41.0
    with an average of 35.4, same as the average of 35.4 in
    \newcite{lewis-etal-2020-bart}.
  \item For XSum, our $\lambda$=none decoding has 45.3/21.9/36.8 with
    an average of 34.7, compared to an average of 34.9 in
    \newcite{lewis-etal-2020-bart}.
  \item For Multi-News, our MN-800 $\lambda$=none decoding has
    50.2/20.5/45.8 with an average of 38.8, compared to improved
    \rouge F$_1$ results of 44.5/16.0/40.3 with an average of 33.6 by
    Fabbri (personal communication) for
    \newcite{fabbri-etal-2019-multi}.
\end{itemize}    

\subsection{Comparing Summarization Models}\label{appendix:models-rouge}

To complement our comparison of different models in
\secref{across-models}, we list the \rouge-L F$_1$ scores of the five
models in \tabref{comparing-systems-rouge}.

\begin{table}[t]
\setlength{\tabcolsep}{9pt}
\centering
\begin{tabular}{@{}llr@{}}
\toprule & Model & RL\\ \midrule

CNN/DM 
& \bart             & 41.0 \\
& \textsc{BertSum}  & 39.2 \\
& \textsc{PGConv}   & 36.4 \\
& \textsc{BottomUp} & 38.3 \\
& \textsc{AbsRL}    & 37.3 \\

\addlinespace[1mm]

XSum 
& \bart             & 36.8 \\ %
& \textsc{BertSum}  & 31.3 \\

\bottomrule
\end{tabular}
\caption{\rouge-L F$_1$ scores for the models compared in
  \secref{across-models}.}\label{tab:comparing-systems-rouge}
\end{table}

\section{Additional Experimental Details}

We used AWS p3.8x and p3.16x EC2 machines for all our experiments,
except we ran FEQA on the Multi-News summaries on a p3dn.24xlarge
machine, as it required more memory.

The \bart model has 406,290,432 parameters. Fine-tuning \bart on the
Multi-News training set took about 2.5 hours on 4 GPUs; we fine-tuned
for 5 epochs following instructions on the fairseq \bart webpage,
without further hyperparameter search. For CNN/DM and XSum we used the
provided checkpoints.\footnote{See
\url{https://github.com/pytorch/fairseq/tree/master/examples/bart}.}
The minimum and maximum length for Multi-News decoding was determined
by the lengths of the training reference summaries.

\end{document}